\documentclass[10pt,twocolumn,letterpaper]{article}

\usepackage{iccv}              
\usepackage{multirow}

\definecolor{iccvblue}{rgb}{0.21,0.49,0.74}
\usepackage[pagebackref,breaklinks,colorlinks,allcolors=iccvblue]{hyperref}

\def\paperID{2711} 
\def\confName{ICCV}
\def\confYear{2025}

\title{SparseRecon: Neural Implicit Surface Reconstruction from Sparse Views with Feature and Depth Consistencies}

\author{Liang Han$^1$, Xu Zhang$^3$, Haichuan Song$^{2*}$, Kanle Shi$^4$, Yu-Shen Liu$^1$\thanks{Corresponding author: Yu-Shen Liu and Haichuan Song.} \ , Zhizhong Han$^5$
\\
$^1$School of Software, Tsinghua University, Beijing, China \\
$^2$Computer Science and Technology, East China Normal University, Shanghai, China \\
$^3$China Telecom \ \ $^4$Kuaishou Technology, Beijing, China \\
$^5$Department of Computer Science, Wayne State University, Detroit, USA \\
{\tt\small hanl23@mails.tsinghua.edu.cn,} \ 
{\tt\small zhangxu@chinatelecom.cn,} \
{\tt\small hcsong@cs.ecnu.edu.cn} \\
{\tt\small shikanle@kuaishou.com,} \ 
{\tt\small liuyushen@tsinghua.edu.cn,} \ 
{\tt\small h312h@wayne.edu}
}

\begin{document}
\maketitle
\begin{abstract}
Surface reconstruction from sparse views aims to reconstruct a 3D shape or scene from few RGB images. 
The latest methods are either generalization-based or overfitting-based. However, the generalization-based methods do not generalize well on views that were unseen during training, 
while the reconstruction quality of overfitting-based methods is
still limited by the limited geometry clues.
To address this issue, we propose SparseRecon, a novel neural implicit reconstruction method for sparse views with volume rendering-based feature consistency and uncertainty-guided depth constraint.
Firstly, we introduce a feature consistency loss across views to constrain the neural implicit field. This design alleviates the ambiguity caused by insufficient consistency information of views and ensures completeness and smoothness in the reconstruction results. Secondly, we employ an uncertainty-guided depth constraint to back up the feature consistency loss in areas with occlusion and insignificant features, which recovers geometry details for better reconstruction quality. Experimental results demonstrate that our method outperforms the state-of-the-art methods, which can produce high-quality geometry with sparse-view input, especially in the scenarios with small overlapping views. Project page: \url{https://hanl2010.github.io/SparseRecon/}.
\end{abstract}    
\section{Introduction}
\label{sec:intro}

As one of the important tasks in computer vision, 3D reconstruction has attracted lots of research attentions in recent years. With the advancement of deep learning, 3D reconstruction based on neural implicit representations learned from point clouds \cite{tian2023superudf, zhou2024capudf, Noise2NoiseMapping, zhou2024noise2noise, noda2025learning} or images \cite{wang2021neus, yariv2021volsdf, liu2024reconclothing, zhou2023hdhuman} becomes a popular research topic.
Although recent multi-view reconstruction methods\cite{wang2021neus, yariv2021volsdf,darmon2022neuralwarp,wang2023neus2,wu2022voxurf, zhang2023fastnerf} have made great progress in terms of the reconstruction quality and reconstruction speed, they require a large number of dense views as supervision.
When limited number of views are available, 
current reconstruction methods usually struggle to reconstruct high-quality surfaces.

\begin{figure}[tb]
    \centering
    \includegraphics[width=1.0\linewidth]{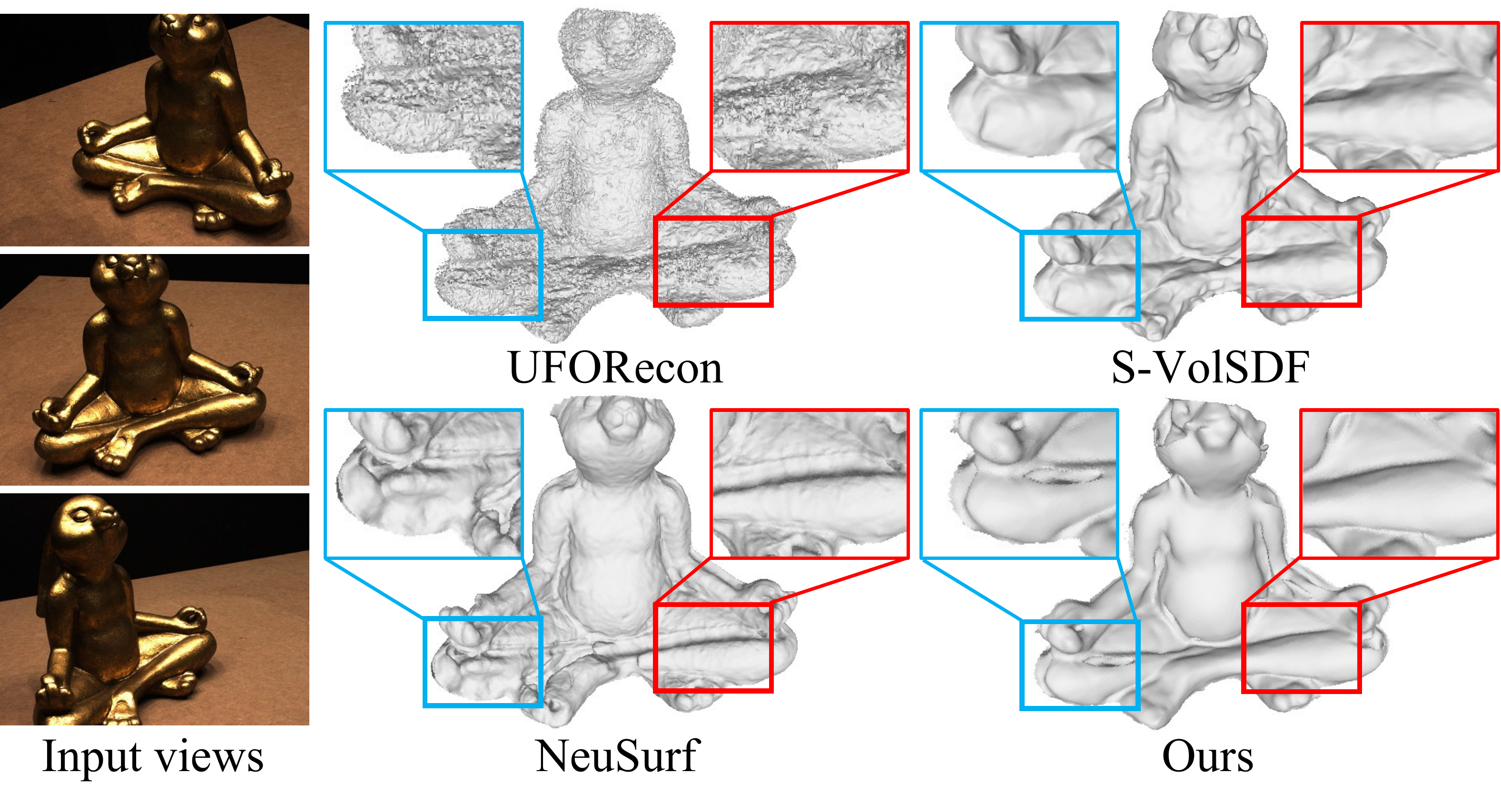}
    \caption{Given only 3 input images with large view angle change, our method can reconstruct a smoother surface compared to the state-of-the-art methods, such as UFORecon \cite{na2024uforecon}, S-VolSDF \cite{wu2023svolsdf} and NeuSurf \cite{huang2023neusurf}.
    The details of each surface are shown in the colored boxes.
    }
    \label{fig:cover}
\end{figure}

Existing methods for sparse view reconstruction can be mainly classified into two categories: generalization-based methods and overfitting-based methods. 
The generalization-based methods \cite{long2022sparseneus, ren2023volrecon, peng2023gens, liang2024retr, na2024uforecon} emphasize the generalization of sparse-view reconstruction, but they are mainly effective in scenarios with large view overlaps. In cases with views that were unseen during training,
the quality of the reconstructed surface degenerates significantly, as shown in Figure \ref{fig:cover}. Meanwhile, it takes a long time to pre-train these methods on large-scale data, but the generalization is still not good enough. Instead, overfitting-based methods \cite{yu2022monosdf, wu2023svolsdf,younes2024sparsecraft, huang2023neusurf, fatesgs} typically fit the 3D geometry directly from the sparse views by leveraging geometry clues. They show promising capability of reconstructing higher-quality geometric surfaces 
with small-overlapping views. However, the reconstruction quality of the existing methods is still unsatisfactory.

In this paper, we introduce a multi-view feature consistency loss based on volume rendering and an uncertainty-guided depth constraint to learn neural signed distance functions. This approach allows us to achieve high-quality mesh reconstruction on more challenging sparse views with small overlaps.

For the feature consistency loss, we first employ the pre-trained Vis-MVSNet \cite{zhang2020vismvsnet} to obtain depth features from the input images. Then, within a neural implicit rendering framework, the sampled 3D points along the rays emitted from the reference image are projected to the source image and the reference image. This allows us to acquire source features and reference features of each 3D point and measure the similarity between these two kinds of features. Finally, the feature similarity for each 3D point along the rays is accumulated through volume rendering, thus yielding the feature similarity associated with the rays. During optimization, we pursue higher feature similarity along the rays. Since the depth information is implicitly encoded with image features, feature consistency constraint can significantly alleviate the ambiguity issues arising from insufficient consistency of sparse views and low-texture during reconstruction. 

For the uncertainty-guided depth prior constraint, we follow MonoSDF \cite{yu2022monosdf}, utilizing a pre-trained network to acquire depth priors for each image, and then use it to constrain the regions with uncertain depth. 
However, monocular depth priors do not have consistent scales to the ground truth depth, which are hard to get calibrated to ground truth either due to the distortion. To effectively leverage the depth priors and provide proper supervision for occluded or under-constrained regions, we propose an uncertainty-guided depth prior constraint. First, we calibrate the depth priors using sparse point clouds obtained from COLMAP \cite{schoenberger2016colmap}. Then, during training, we compute the depth confidence from the rendered depth and impose the depth prior constraint only in regions with low confidence. This constraint helps infer more accurate geometry in occluded or under-constrained regions, minimizing the negative impact of depth prior errors on well-constrained regions.

We evaluate our methods on several widely used benchmarks and report the state-of-the-art results.
In summary, our main contributions are as follows.
\begin{itemize}
    \item We propose a novel feature consistency loss based on volume rendering. It can effectively constrain the neural radiance field by leveraging feature consistency among multiple views, improving the performance in sparse-view reconstruction tasks.
    \item By incorporating depth confidence, we utilize the calibrated depth prior more effectively to enhance geometric constraints, further improving the reconstruction quality.
    \item Extensive experiments on the well-known datasets, such as DTU \cite{jensen2014dtu} and BlendedMVS\cite{yao2020blendedmvs}, demonstrate that our method outperforms existing sparse-view reconstruction methods and achieves the state-of-the-art results.
\end{itemize}

\begin{figure*}[!htb]
    \centering
    \includegraphics[width=0.9\textwidth]{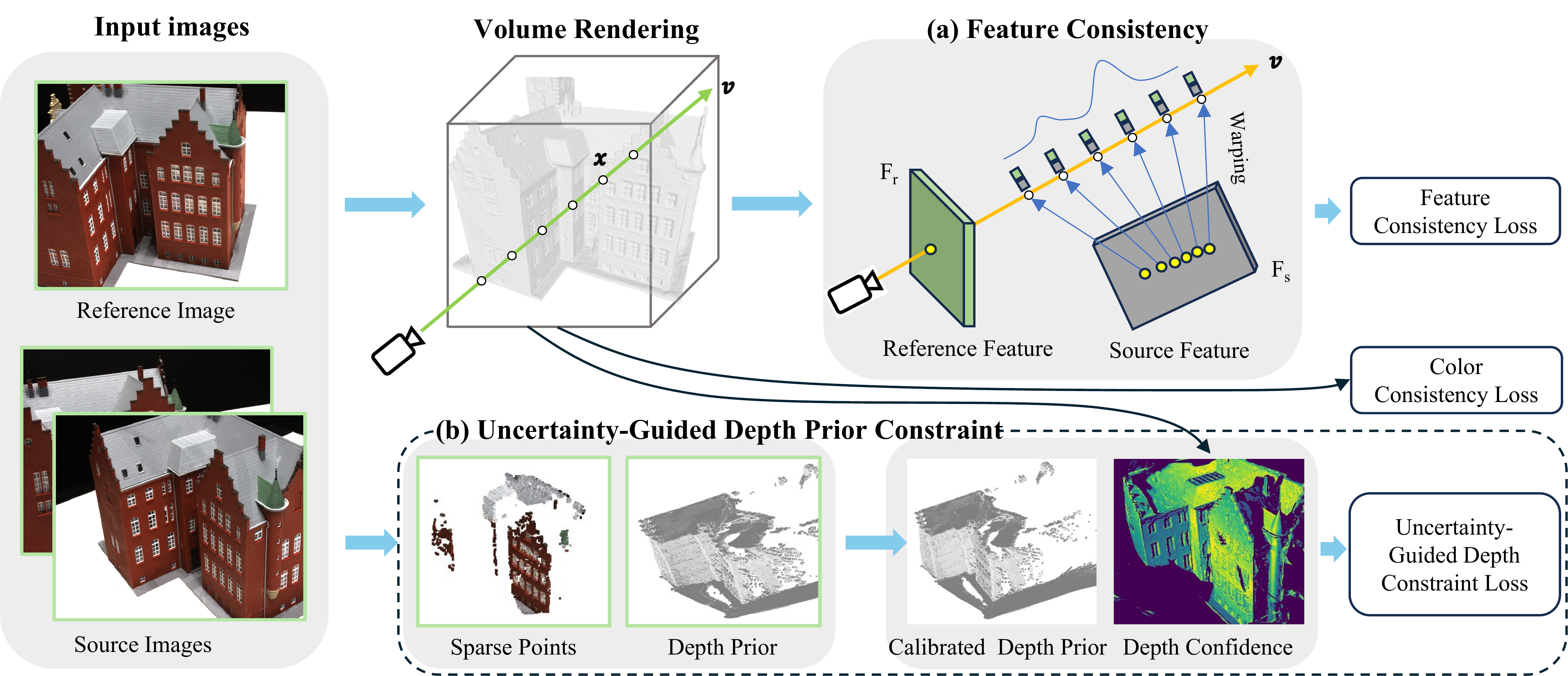}
    \caption{SparseRecon consists of two main parts.
    (a) Volume rendering-based feature consistency constraint. We extract features from the reference image and source images. For a ray emitted from the reference image, we project each sampled point on the ray onto the source images to obtain the corresponding features. Then, the volume rendering-based feature consistency loss is calculated using the corresponding features on the reference image.
    (b) Uncertainty-guided depth prior constraint. We use another pre-trained network to obtain the depth prior of the reference image and calibrate it with the sparse point cloud obtained by COLMAP. Then, we calculate the confidence of the rendered depth, so that the calibrated depth prior only constrains areas with low confidence.
    }
    \label{fig:method_pipeline}
\end{figure*}

\section{Related Work}
\subsection{Neural Implicit Reconstruction}
Neural implicit reconstruction methods \cite{wang2021neus,darmon2022neuralwarp,yu2022monosdf,fu2022geoneus,wu2022voxurf,li2023neuralangelo, zhang2025nerfprior, zhang2025monoinstance} have been rapidly developed based on neural volume rendering \cite{mildenhall2021nerf}. These methods introduce the Signed Distance Function (SDF) as the implicit representation of 3D surfaces in volume rendering in multi-view 3D reconstruction. While these methods have made significant improvements in both reconstruction quality and speed, it is important to note that they heavily rely on multiple views with large overlaps during the optimization.

\noindent\textbf{Generalization-based surface reconstruction with sparse views.}
In order to directly generalize the reconstruction results on sparse views, methods \cite{long2022sparseneus, ren2023volrecon, peng2023gens, xu2023c2f2neus, liang2024retr, na2024uforecon} adopt the strategy of aggregating features from multiple view images to construct a feature volume, which is then used to predict the SDF for reconstructing the surface. VolRecon \cite{ren2023volrecon} uses transformers \cite{katharopoulos2020transformers} to aggregate multi-view features, C2F2NeUS \cite{xu2023c2f2neus} employs cascade architecture to construct a volume pyramid, while ReTR \cite{liang2024retr} and UFORecon \cite{na2024uforecon} aggregates multi-level features. These methods require pretraining on large-scale datasets, which typically takes several days. However, when there is a significant domain gap between the testing and training data, they all fail to reconstruct shapes effectively.


\noindent\textbf{Overfitting-based surface reconstruction with sparse views.} 
In contrast, overfitting-based methods directly fit the 3D geometry from the sparse images by geometric prior constraints.
MonoSDF \cite{yu2022monosdf} employs depth and normal priors to achieve sparse reconstruction with small-overlapping views. However, such priors come with errors, and it does not fully leverage inter-view consistency, resulting in lower reconstruction quality. 
S-VolSDF \cite{wu2023svolsdf} employs probability volumes obtained from MVS \cite{gu2020cascademvs} models to guide the rendering weight estimated by VolSDF \cite{yariv2021volsdf}. This improves the reconstruction results in sparse views with small overlap. However, the uncertainties in volumes make negative impact on the reconstructed surface, leading to surface roughness or significant defects. 
NeuSurf \cite{huang2023neusurf} leverages sparse point clouds and employs CAP-UDF \cite{zhou2024capudf} to construct an implicit geometric prior to improve the reconstruction quality from sparse views. However, when the sparse point cloud fails to cover the majority of the object surface, implicit geometric prior information cannot be useful. In contrast, our method employs more robust feature priors, calculates feature consistency based on volume rendering, and simultaneously utilizes depth priors to optimize the occluded regions, ultimately resulting in high-quality surface.

\subsection{Gaussian Splatting} 
Gaussian Splatting \cite{gaussian} has achieved unprecedented optimization speed and rendering quality in the task of novel view synthesis \cite{li2024dngaussian, zhu2025fsgs, han2024binocular, gap}. However, since the Gaussians are unorganized, the discrete and unstructured points make it difficult to extract 3D surfaces through post-processing. To address this issue, some methods introduce to use regularization terms \cite{guedon2023sugar}, convert 3D Gaussians to 2D surfels \cite{huang20242dgs, dai2024gs_surfel}, acquire opacity fields through rays \cite{yu2024gof}, improve the depth rendering algorithm \cite{pgsr} of 3DGS, or jointly optimize 3DGS with neural radiance fields \cite{lyu20243dgsr, chen2023neusg, zhang2024gspull, li2025gaussianudf}.
However, these methods require dense views to work well.
Recently, FatesGS \cite{fatesgs} achieves fast sparse-view reconstruction by leveraging depth priors and on-surface feature consistency constraints. 
Nevertheless, the reconstruction results still exhibit roughness or noticeable defects.

\subsection{Sparse View Synthesis}
In addition, the novel view synthesis from sparse views is another category of work closely related to sparse view reconstruction. 
Depending on the technical framework, these works can be categorized into NeRF-based methods \cite{truong2023sparf, wang2023sparsenerf, yang2023freenerf, niemeyer2022regnerf, jain2021dietnerf, yuan2022rgbdnerf, deng2022dsnerf} and Gaussian Splatting-based methods \cite{li2024dngaussian, zhu2025fsgs, han2024binocular, chen2025mvsplat, zhang2024corgs}.
This line of research also employs a limited number of views as input. However, they solely focus on the rendering quality of novel views rather than surface reconstruction, which are not designed specifically for the accurate geometric surface reconstruction. 
Due to the discernible bias (i.e. inherent geometric errors) \cite{wang2021neus} caused by the conventional volume rendering or inconsistencies in depth that appear in Gaussian rendering, current sparse view synthesis methods still fail to correctly reconstruct high-fidelity geometric surfaces.

\section{Method}
\label{sec:methods}
The overview of our method is depicted in Figure~\ref{fig:method_pipeline}. We introduce a novel feature consistency loss and an uncertainty-guided depth constraint based on the NeuS \cite{wang2021neus} framework. In this section, 
we first explain how to compute feature consistency for sampled points along rays. Then we explain how to enhance geometric constraints using depth priors and depth uncertainty. Thirdly, we introduce the color consistency loss. Finally, we present the overall loss function for optimization.



\subsection{Volume Rendering-based Feature Consistency}
\label{subsec:feat_loss}
First, we use a pre-trained MVS network \cite{zhang2020vismvsnet} to extract the features from both the reference image and the source image. Given a ray emitted from the reference image, let $p_r(0)$ denote the point where a ray intersects the reference image. And for each point $x_i$ along the ray, we denote its projection on the source image as $p_s(i)$. Then, we bilinearly interpolate $F_r(0)$ and $F_s(i)$ at points $p_r(0)$ and $p_s(i)$ on image features, respectively. Formally, we define the feature consistency loss function as follows,
\begin{equation}
    L_{feat}=M^{occ}(1-\frac{1}{N}\sum_{i=1}^{N}w_if_{cos}(F_r(p_r(0)),F_s(p_s(i))), 
    \label{eq: feat_loss}
\end{equation}
where $f_{cos}$ is the cosine similarity, and $w_i$ corresponds to the weight for each point along the ray. $p_s(i)=K(Rx_i + t)$ is the projection of $x_i$ in source view, and $[K; R; t]$ is the camera parameters of source view. $M^{occ}$ is the occlusion mask.

Although MVSDF \cite{zhang2021mvsdf}, NeuSurf \cite{huang2023neusurf} and FatesGS \cite{fatesgs} also employ feature consistency constraints, they just leverage the intersection point between a camera ray and the object's surface. Then, this intersection point gets projected onto adjacent views to obtain the corresponding image features for the purpose of comparing features at this point across multiple views. In sparse view scenarios, the estimated positions of surface points can easily deviate significantly, making the on-surface feature consistency loss not converge. NeuSurf \cite{huang2023neusurf} and FatesGS \cite{fatesgs} utilize sparse point clouds generated by COLMAP \cite{schoenberger2016colmap} as priors, enabling it to obtain partially accurate positions of surface points, thereby allowing the on-surface feature consistency loss to be more effectively leveraged. However, in regions of lacking surface points, the on-surface feature consistency loss cannot ensure the attainment of high-quality geometric surfaces.

\begin{figure}[t]
    \centering
    \includegraphics[width=1.0\linewidth]{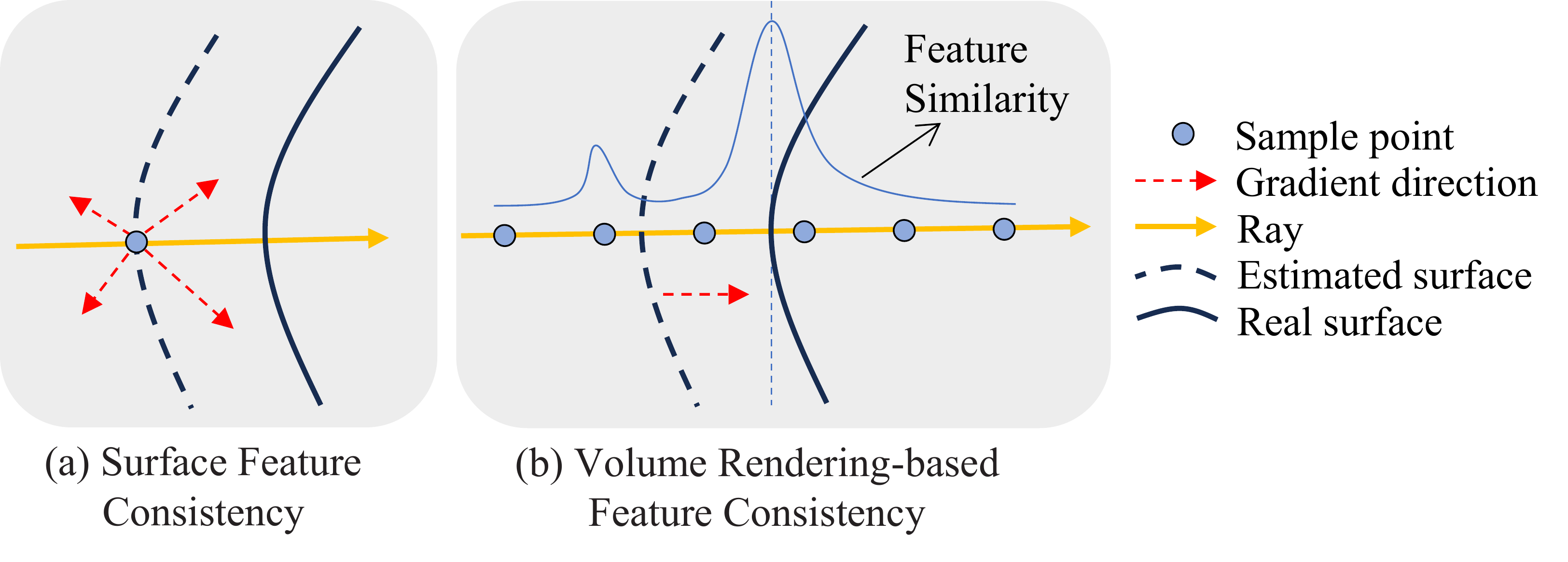}
    \caption{The illustration of (a) on-surface feature consistency and (b) feature consistency with volume rendering.}
    \label{fig:feat_consis}
\end{figure}

Figure \ref{fig:feat_consis} illustrates the difference between on-surface feature consistency and volume rendering-based feature consistency. Due to the uncertainty of gradient direction, the constraint solely relying on surface point features is challenging to be optimized.
In contrast, our method does not require the prior estimation of surface points, it calculates feature consistency on all sampling points along the ray, and provides more reasonable and comprehensive supervision to the implicit field, thereby addressing the convergence issue that may arise in sparse reconstruction for MVSDF \cite{zhang2021mvsdf} and NeuSurf \cite{huang2023neusurf}.

\subsection{Uncertainty-Guided Depth Constraint}
\label{subsec:depth_loss}

\begin{figure}
    \centering
    \includegraphics[width=1.0\linewidth]{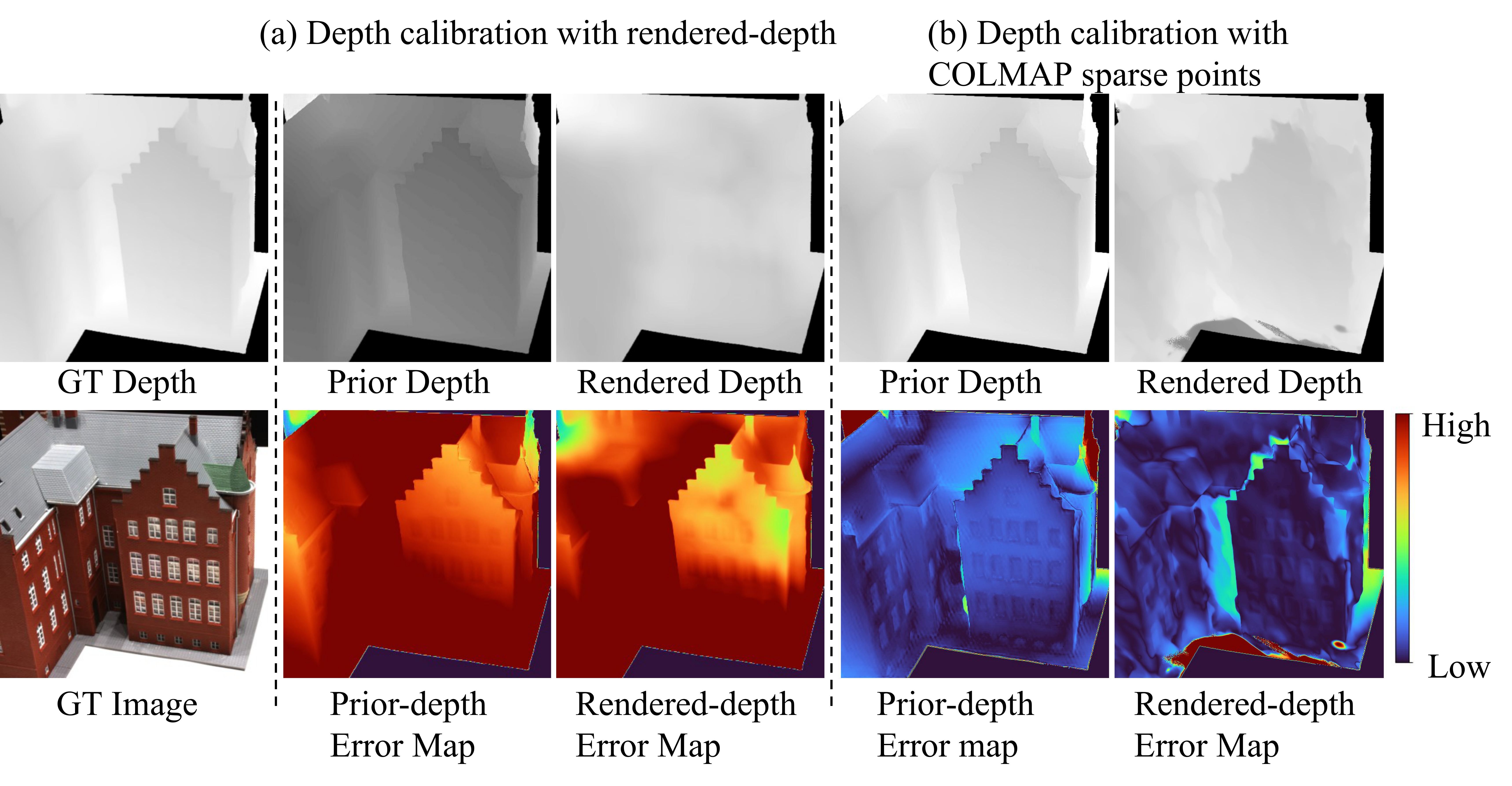}
    \caption{The illustration of rendered depth produced by different depth prior utilization methods, along with the corresponding error maps. (a) Calibrate the depth prior using the rendered depth during training. (b) Calibrate the depth prior using the COLMAP sparse point cloud.}
    \label{fig:depth_comparision}
\end{figure}

Although multi-view features offer more robust constraint than image colors, they are ineffective for occluded regions. Due to the limited number of views, some regions may only be visible from a single viewpoint. To enhance geometric constraints, we employ depth priors to supervise the radiance field. However, monocular depth priors are not perfect and accurate. Although MonoSDF \cite{yu2022monosdf} has already taken the inaccuracy of depth priors into account, i.e., it aligns depth priors using rendered depth during training. However, the rendered depth during training is inaccurate, resulting in significant errors in the calibrated depth priors. This ultimately leads to the accumulation of errors during training, which results in inaccurate reconstructions. Figure \ref{fig:depth_comparision} (a) shows the calibrated depth prior and rendered depth obtained by MonoSDF \cite{yu2022monosdf}, as well as their error maps compared to the ground truth depth. It can be seen that both the calibrated depth prior and the rendered depth are with large errors. Therefore, MonoSDF \cite{yu2022monosdf} uses a weight annealing stategy to anneal the weight of depth loss to 0 during the first 200 training epochs.

Another trivial approach is to calibrate the depth priors using the sparse point cloud obtained from COLMAP \cite{schoenberger2016colmap}. Since the sparse points are generally located on the geometric surface of the object, their depth is relatively accurate. Therefore, calibrating the depth priors using the sparse point cloud can lead to more accurate depth priors. Figure \ref{fig:depth_comparision} (b) shows the depth priors calibrated with the sparse point cloud, and the rendered depth, as well as their error maps compared to the ground truth depth. It indicates that the depth priors calibrated to the point cloud from the COLMAP \cite{schoenberger2016colmap} are more accurate. Therefore, we can use them as an constraint leads to more precise rendered depth.

However, due to the distortions in monocular depth priors, it is impossible to perfectly align them with the ground truth depth. Even after calibration, the depth priors still exhibit noticeable errors when compared to the ground truth depth. In sparse view scenarios, occlusions and insufficient constraints are more common, leading to significant discrepancies between the geometry of occluded regions and the real surface. Therefore, to achieve more accurate geometry in these under-constrained regions while avoiding the negative impact of depth prior errors on well-constrained regions, we propose an uncertainty-guided depth prior constraint method to more effectively utilize the depth priors. Specifically, we apply depth prior constraints in regions with depth uncertainty, while refraining from using them in regions with high depth confidence.

\begin{figure}
    \centering
    \includegraphics[width=1.0\linewidth]{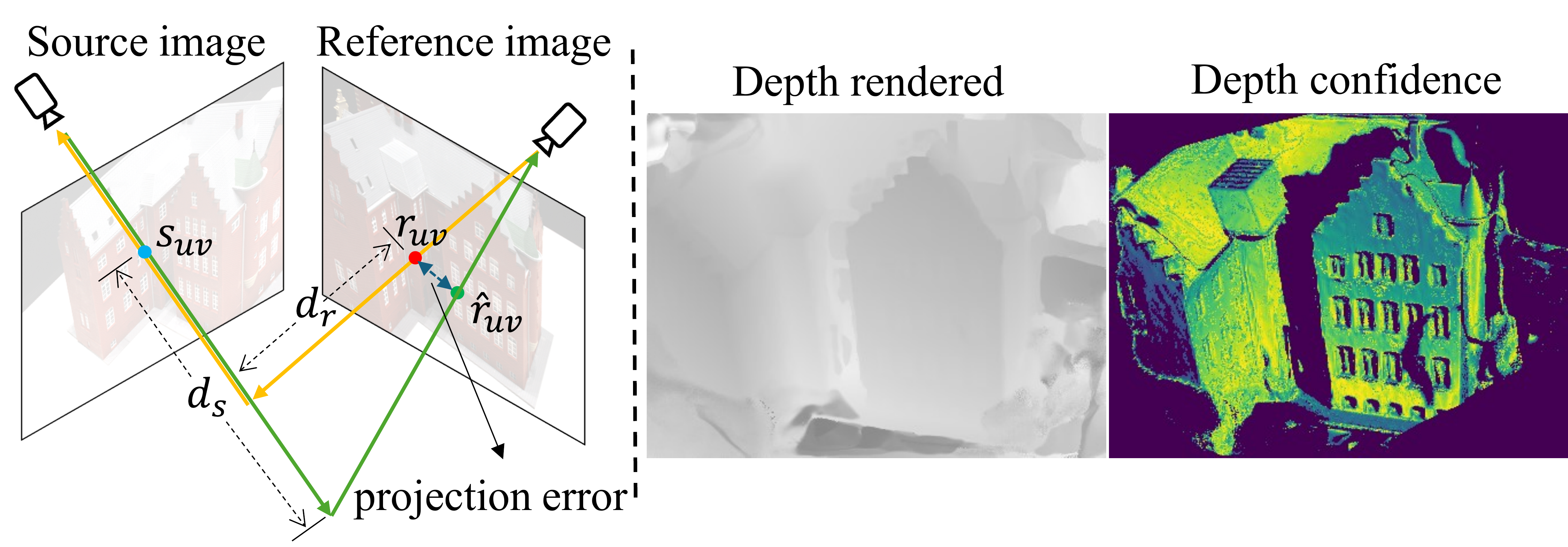}
    \caption{Left: the method of obtaining the confidence of rendered depth. Right: the rendered depth and the depth confidence.}
    \label{fig:depth_confidence}
\end{figure}

To obtain the confidence of the rendered depth, 
we employ a method to evaluate the multi-view depth projection consistency. As shown in Figure \ref{fig:depth_confidence}, for a specific pixel $r_{uv}$ in the reference image with depth $d_r$, it can be mapped to a neighboring image through the homography matrix $H_{rs}$, leading to a pixel $s_{uv}$,

\begin{equation}
    s_{uv} = H_{rs} r_{uv},
\end{equation}
\begin{equation}
    H_{rs} = M_s M_r^{-1},
\end{equation}

\noindent where $M_r$ and $M_s$ are the projection matrices corresponding to the reference and source views, respectively. Similarly, we can map the pixel $s_{uv}$ in the source view to the reference view using the projection matrix $H_{sr}$ and its corresponding depth $d_s$, resulting in $\hat{r}_{uv}$. The forward and backward projection distance error reflect the accuracy of depth predictions, so we take it as the depth confidence, which is defined as

\begin{equation}
    C_d = 
    \begin{cases}
    \frac{1}{e^{\left \| r_{uv} - \hat{r}_{uv} \right \| }}, &\text{ if } \left \| r_{uv} - \hat{r}_{uv} \right \| \leqslant 1 \\
    0, &\text{ if } \left \| r_{uv} - \hat{r}_{uv} \right \| >1
    \end{cases}
\end{equation}

The right side of Figure \ref{fig:depth_confidence} shows the rendered depth and the corresponding depth confidence.

Correspondingly, the depth uncertainty is defined as $U_d=1-C_d$. Meanwhile, we can set a threshold $\tau$ for depth confidence $C_d$ to obtain the occlusion mask $M^{occ}=\{C_d>\tau\}$.

For depth calibration, we leverage COLMAP \cite{schoenberger2016colmap} to obtain a sparse point cloud $\{X:x_1,x_2 \dots x_i\in R\}$ and visibility flags indicating which keypoints are visible from view $I$. Given the camera parameters $P$ of view $I$, we estimate the depth $\bar{D_i}$ of keypoints by computing the distance from the visible keypoints $x_i$ to the camera center $o$. Then, we calibrate the monocular depth prior $\hat{D}$ with $\bar{D_i}$, it can be defined as $\bar{D} \approx a\hat{D} + b$, where $a$ is the scale factor and $b$ is the shift factor, obtained through the least squares method. 
Formally, the depth constraint loss is defined as,
\begin{equation}
    L_{depth} = \sum_{r\in R}^{} U_d \left \| (a\hat D+b) - D_{pred} \right \| ^2. 
    \label{eq:depth_loss}
\end{equation}

\subsection{Color Consistency Constraint}
\label{subsec:color_loss}
Although feature consistency constraint can ensure that the reconstruction does not suffer from severe artifacts, it does not provide sufficient supervision to reconstruct fine geometric details. Conversely, in cases with rich textures, image color constraint can refine the geometric details. Therefore, following the NeuralWarp \cite{darmon2022neuralwarp}, pixel warping loss and patch warping loss are used in our method as multi-view color consistency loss functions,
\begin{equation}
\begin{split}
    L_{color}=\sum_{r\in R} M^{occ} d_{pixel}(I(r),I_s(r))  \\ 
    + \sum_{r\in R}M^{occ} d_{patch}(P(r),P_s(r)),
\end{split}
\label{eq:color_loss}
\end{equation}
where $I(r)$ and $I_s(r)$ are the ground truth color of the pixel from which the ray emits and the rendered color, respectively, $P(r)$ and $P_s(r)$ are the ground truth color of the patch corresponding to the ray and the rendered patch color, respectively. $d_{pixel}$ is the loss metric for pixel color, where we use $L1$ loss as $d_{pixel}$. $d_{patch}$ is the loss metric for patch color, where we use the Structural Similarity Index Measure (SSIM \cite{wang2004ssim}) as $d_{patch}$.



\subsection{Training Loss}
\label{subsec:loss_all}
In addition to the above-mentioned three loss functions, we also use the Eikonal loss \cite{gropp2020Eikonal} used in NeuS \cite{wang2021neus}. We define the overall loss function as follows:
\begin{equation}
    L = L_{feat} + \alpha L_{depth} + L_{color} + \beta L_{eik},
    \label{eq:total_loss}
\end{equation}

%
$L_{eki}$ is the Eikonal loss \cite{gropp2020Eikonal}, used to regularize the SDF values of sampled points, defined as
\begin{equation}
    L_{eki.} = \frac{1}{mn}\sum_{i,k}^{} (\left \| \nabla f(x_{i,k}) \right \|_2 -1)^2.
    \label{eq:eki_loss}
\end{equation}
\section{Experiments}

\begin{table*}[t]
  \centering
  \begin{tabular}{@{}lcccccccccccc@{}}
    \toprule
    Methods & 21 & 24 & 34 & 37 & 38 & 40 & 82 & 106 & 110 & 114 & 118 & Mean CD $\downarrow$ \\
    \midrule
    NeuS \cite{wang2021neus} & 5.63 & 3.58 & 6.00 & 4.60 & 2.57 & 4.53 & 1.91 & 4.18 & 5.46 & 1.19 & 4.16 & 3.98 \\
    NeuralWarp \cite{darmon2022neuralwarp} & 2.53 & 1.88 & \underline{0.74} & \underline{1.80} & \textbf{0.84} & 11.50 & 2.64 & 2.10 & 4.37 & 1.19 & 2.63 & 2.93 \\
    MonoSDF \cite{yu2022monosdf} & 4.14 & 5.92 & 1.39 & 4.55 & 2.19 & 2.14 & 2.36 & 5.62 & 4.58 & 1.63 & 3.02 & 3.41 \\
    Vis-MVSNet \cite{zhang2020vismvsnet} & 3.39 & 4.44 & 0.85 & 3.36 & 1.69 & 3.35 & 3.35 & 2.34 & 2.16 & 0.74 & 1.83 & 2.50 \\
    MVSDF \cite{zhang2021mvsdf} & 4.31 & 4.71 & 1.65 & 6.37 & 1.77 & 4.47 & 3.61 & 1.87 & 1.67 & 1.25 & 1.69 & 3.03 \\
    \hline
    SparseNeuS$_{ft}$ \cite{long2022sparseneus} & 3.48 & 4.37 & 2.92 & 4.76 & 2.79 & 3.73 & 2.80 & 1.86 & 3.10 & 1.15 & 2.29 & 3.02 \\
    VolRecon \cite{ren2023volrecon} & 2.72 & 3.07 & 1.82 & 4.32 & 2.14 & 3.04 & 3.00 & 2.56 & 2.81 & 1.49 & 3.22 & 2.75 \\
    GenS$_{ft}$ \cite{peng2023gens} & 5.86 & 7.67 & 3.62 & 8.57 & 5.37 & 5.41 & 5.48 & 6.04 & 5.29 & 4.69 & 4.35 & 5.67 \\
    ReTR \cite{liang2024retr} & 2.67 & 3.37 & 1.62 & 3.68 & 1.87 & 3.40 & 3.67 & 2.84 & 2.85 & 1.56 & 2.35 & 2.72 \\
    UFORecon \cite{na2024uforecon} & \textbf{1.84} & 1.52 & 0.79 & 2.58 & 1.00 & \underline{1.82} & \underline{1.72} & 1.20 & 0.93 & 0.66 & 1.26 & \underline{1.39} \\
    \hline
    S-VolSDF \cite{wu2023svolsdf} & 2.45 & 3.08 & 1.33 & 3.09 & 1.22 & 3.21 & 1.91 & 1.51 & 1.23 & 0.74 & 1.2 & 1.91 \\
    SparseCraft \cite{younes2024sparsecraft} & 2.88 & 2.42 & 0.92 & 2.97 & 1.58 & 2.78 & 2.51 & 1.10 & 5.24 & 0.65 & 0.88 & 2.16 \\
    NeuSurf \cite{huang2023neusurf} & 7.60 & 1.43 & 2.93 & 3.18 & 1.53 & 2.86 & 1.86 & \underline{1.09} & 1.41 & \textbf{0.37} & \textbf{0.62} & 2.26 \\
    FatesGS \cite{fatesgs} & 3.98 & \underline{1.32} & 2.53 & 2.85 & 3.36 & 2.71 & 3.76 & 1.49 & \underline{0.85} & 0.47 & 1.06 & 2.22\\
    Ours & \underline{2.14} & \textbf{1.26} & \textbf{0.72} & \textbf{1.46} & \underline{0.86} & \textbf{1.39} & \textbf{1.37} & \textbf{0.94} & \textbf{0.77} & \underline{0.44} & \underline{0.83} & \textbf{1.11} \\
    \bottomrule
  \end{tabular}
  \caption{Quantitative results of  Chamfer Distance (CD$\downarrow$) on DTU dataset with \textit{3 small-overlapping} images. The methods are divided into three categories, from top to bottom: (1) dense-view reconstruction methods related to ours, (2) generalization-based sparse-view reconstruction methods, and (3) overfitting-based sparse-view reconstruction methods. the best results are in \textit{bold}, the second best are \textit{underlined}.}
  \label{tab:cd_small_overlap}
\end{table*}

\subsection{Dataset}
We evaluate our method on DTU \cite{jensen2014dtu} and BlendedMVS \cite{yao2020blendedmvs} dataset.
For the DTU \cite{jensen2014dtu} dataset, to avoid using the scenes that have already been used as training data on the pretrained Vis-MVSNet \cite{zhang2020vismvsnet} model, we select the same 11 scenes as in S-VolSDF \cite{wu2023svolsdf}. The image resolution is set to 1600$\times$1200. 
Similar to the S-VolSDF \cite{wu2023svolsdf} and NeuSurf \cite{huang2023neusurf} methods, we select the views 22, 25, and 28 for the more challenging reconstruction of small overlaps.

For the BlendedMVS \cite{yao2020blendedmvs} dataset, we follow the S-VolSDF \cite{wu2023svolsdf} to use the same 9 challenging scenes, with 3 small-overlapping views for each scene.
The image resolution is set to 768$\times$576.


\subsection{Implementation Details}
We use the same network architecture and initialization strategy as NeuS \cite{wang2021neus} and incorporated our volume rendering feature consistency loss, uncertainty-guided depth constraint loss, and color consistency loss. For the weight factors in the loss functions Eq.~\ref{eq:total_loss}, we set the $\alpha$ for the uncertainty-guided depth prior constraint loss $L_{depth}$ to 0.5 and the $\beta$ for the Eikonal loss $L_{eik}$ to 0.1. Each scene is trained 100K iterations on a RTX3090 GPU. The patch warping term in the color consistency loss requires the surface point normals to calculate homographies, but the initial normals are too noisy \cite{darmon2022neuralwarp}, therefore, the patch warping loss is applied after 20k training steps. The threshold $\tau$ of the occlusion mask is set to 0. 

\subsection{Baseline}
We compare our approach with three categories of methods including \textit{dense-view methods}: NeuS \cite{wang2021neus},
NeuralWarp \cite{darmon2022neuralwarp}, Vis-MVSNet \cite{zhang2020vismvsnet}, MVSDF \cite{zhang2021mvsdf},
\textit{generalization-based methods}: SparseNeuS \cite{long2022sparseneus}, VolRecon \cite{ren2023volrecon}, GenS \cite{peng2023gens}, ReTR \cite{liang2024retr} and UFORecon \cite{na2024uforecon}, \textit{overfitting-based methods}: S-VolSDF \cite{wu2023svolsdf}, SparseCraft \cite{younes2024sparsecraft}, NeuSurf \cite{huang2023neusurf} and FatesGS \cite{fatesgs}. The reconstruction results for SparseNeuS \cite{long2022sparseneus} and GenS \cite{peng2023gens} are fine-tuned using 3 views for each scene.



\subsection{Comparisons}


\begin{figure*}[!htb]
    \centering
    \includegraphics[width=1.0\textwidth]{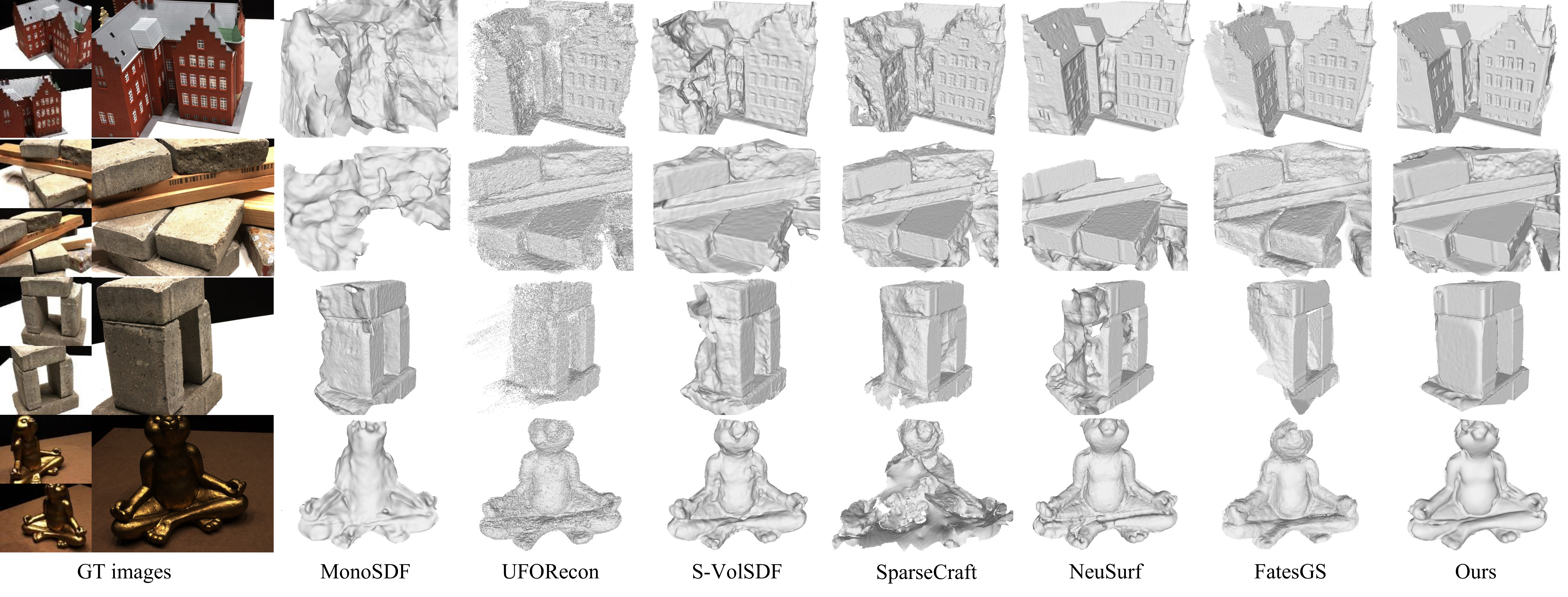}
    \caption{Visual comparison on DTU dataset with \textit{3 small-overlapping} images.}
    \label{fig:dtu_small_overlap}
\end{figure*}

\paragraph{Reconstruction on DTU.} 
For a comprehensive comparison, we evaluate the baselines and our method on both small-overlapping and large-overlapping views. Following baselines \cite{long2022sparseneus, huang2023neusurf, fatesgs}, we report the Chamfer Distance (CD) between the reconstruction surfaces and the ground truth point clouds. Since the scenes in our experiments differ from those used in NeuSurf \cite{huang2023neusurf} and FatesGS \cite{fatesgs}, we re-ran the official implementations of NeuSurf and FatesGS. 
The CD results with small overlapping views are shown in Table~\ref{tab:cd_small_overlap}. The meshes reconstructed by several methods using 3 views with small overlapping are shown in \cref{fig:dtu_small_overlap}. For the generalization-based sparse reconstruction methods, we only show the reconstruction results of the latest UFORecon \cite{na2024uforecon}, as the reconstruction quality of other methods is lower than that of UFORecon \cite{na2024uforecon}. The experimental results show that our method significantly improves the mesh quality with small overlap views, compared to the state-of-the-art sparse-view reconstruction methods. The results of large overlapping views are presented in the supplementary materials.

As shown in Figure~\ref{fig:dtu_small_overlap}, when input sparse views with small overlap, both MonoSDF \cite{yu2022monosdf} and SparseCraft \cite{younes2024sparsecraft} suffer from reconstruction ambiguity and failures, highlighting that relying solely on simplistic geometric prior constraints is insufficient to obtain complete and accurate meshes. UFORecon \cite{na2024uforecon} shows significant roughness in its reconstruction results. S-VolSDF \cite{wu2023svolsdf}, NeuSurf \cite{huang2023neusurf} and FatesGS \cite{fatesgs} exhibit noticeable reconstruction defects. Experimental results demonstrate that our method is effective in alleviating geometric and appearance ambiguities during the optimization process. This significantly enhances the quality of mesh reconstruction, especially in scenarios with small overlapping views and low texture.


\begin{figure}[htb]
    \centering
    \includegraphics[width=1.0\linewidth]{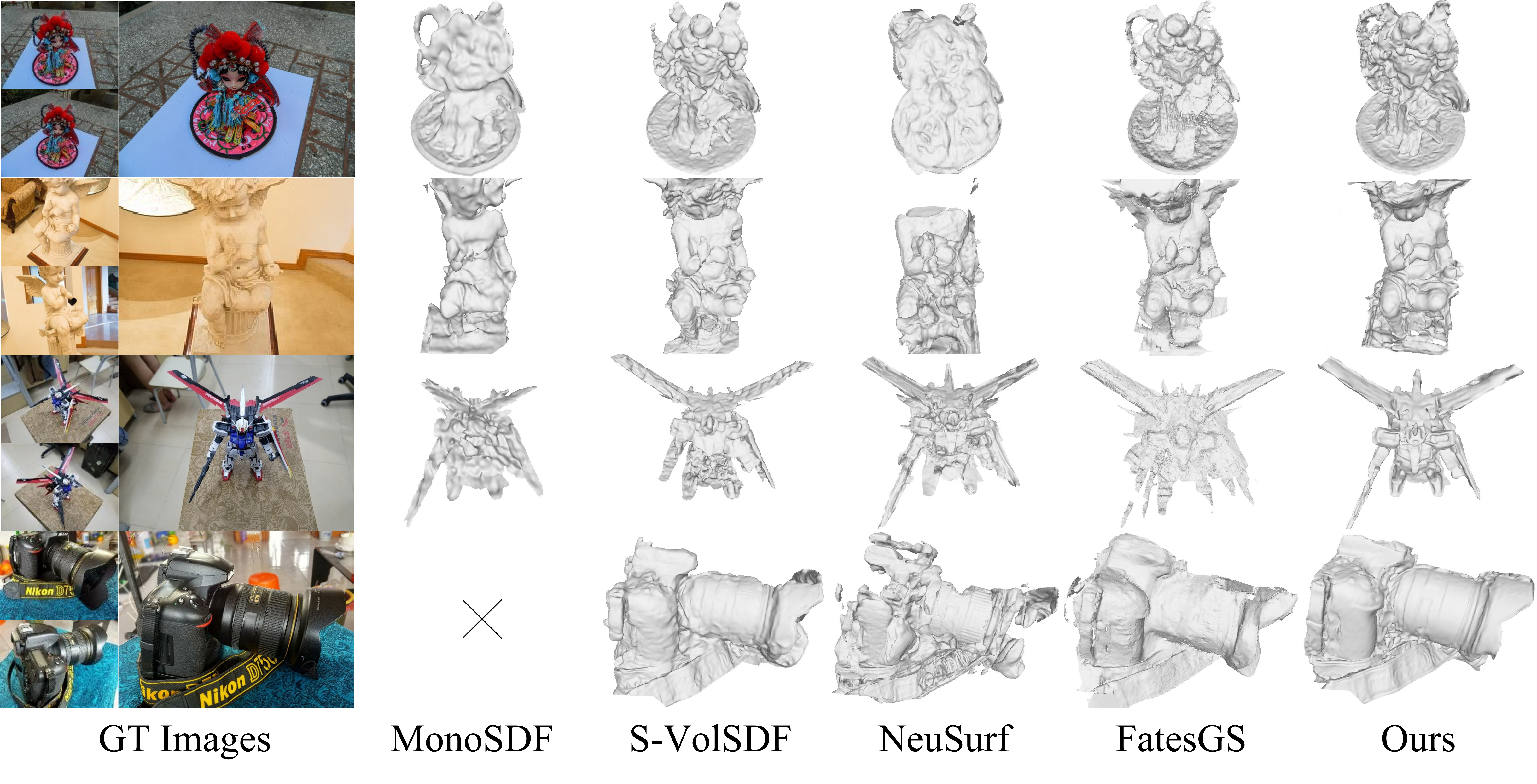}
    \caption{Visual comparison on BlendedMVS dataset. '$\times$' indicates reconstruction failure.}
    \label{fig:bmvs_three_image}
\end{figure}

\paragraph{Reconstruction on BlendedMVS.}
Figure~\ref{fig:bmvs_three_image} presents the visual comparison of reconstructed mesh for overfitting-based
methods. With only 3 small-overlapping views provided, all of the generalization-based methods completely fail to reconstruct in the sparse setting of BlendedMVS\cite{yao2020blendedmvs} dataset, even if SparseNeuS \cite{long2022sparseneus} is fine-tuned. Therefore, the reconstruction results of these methods are not included in Figure~\ref{fig:bmvs_three_image}. 
Compared to other methods, our approach can generate more complete and detailed meshes. Similarly, MonoSDF \cite{yu2022monosdf} fails to reconstruct surfaces either. The meshes generated by S-VolSDF \cite{wu2023svolsdf}, NeuSurf \cite{huang2023neusurf} and FatesGS \cite{fatesgs} exhibit significant defects. Both NeurSurf \cite{huang2023neusurf} and FatesGS \cite{fatesgs} use on-surface feature consistency constraints, but the reconstruction results are still not good enough. In contrast, our method achieves more comprehensive geometry and finer details by employing volume rendering-based feature consistency constraints. This highlights the advantages of our approach in geometric consistency. More visualizations are presented in the supplementary materials.

\begin{figure}[htb]
    \centering
    \includegraphics[width=1.0\linewidth]{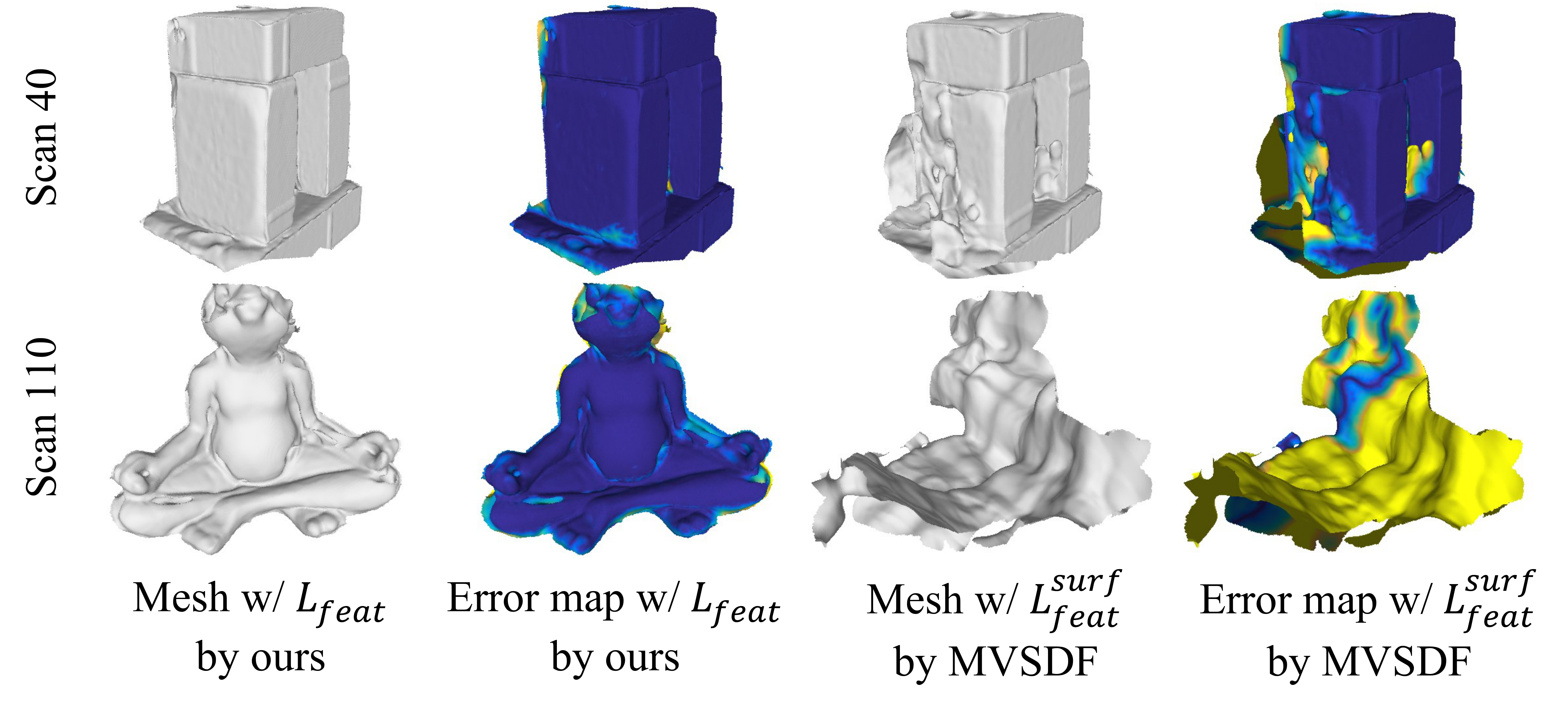}
    \caption{Reconstructed meshes and error maps on DTU dataset with different feature consistency losses. }
    \label{fig:surf_feat_and_ray_feat}
\end{figure}

\begin{table}[]
    \centering
    \resizebox{\linewidth}{!}{
    \begin{tabular}{c|ccccccc|c}
      \hline
      Method & $L_{color}$ & $L_{feat}$ & $L_{depth}$ & $L_{depth}^{mono}$ & $L_{feat}^{L1}$ & $L_{feat}^{L2}$ & $L_{feat}^{surf}$ & CD$\downarrow$ \\
      \hline
      Baseline   &            &            &                &             & & & & 3.35 \\
               &  \checkmark  &            &                &             & & & & 1.76 \\
               &  \checkmark  & \checkmark &                &             & & & & 1.47 \\
               &  \checkmark  &            & \checkmark     &             & & & & 1.62 \\
               & \checkmark   & \checkmark & \checkmark     &             & & & & \textbf{1.11} \\
               & \checkmark   & \checkmark &                & \checkmark  & & & & 1.59 \\
               & \checkmark   &            & \checkmark     &  & \checkmark & & & 2.36 \\
               & \checkmark   &            & \checkmark     &  & & \checkmark & & 1.81 \\
               & \checkmark   &            & \checkmark     &  & & & \checkmark & 2.93 \\
      \hline
    \end{tabular}}
    \caption{Ablation studies on DTU dataset with 3 small-overlapping images.}
    \label{tab:ablation}
\end{table}

\begin{figure}[htb]
    \centering
    \includegraphics[width=1.0\linewidth]{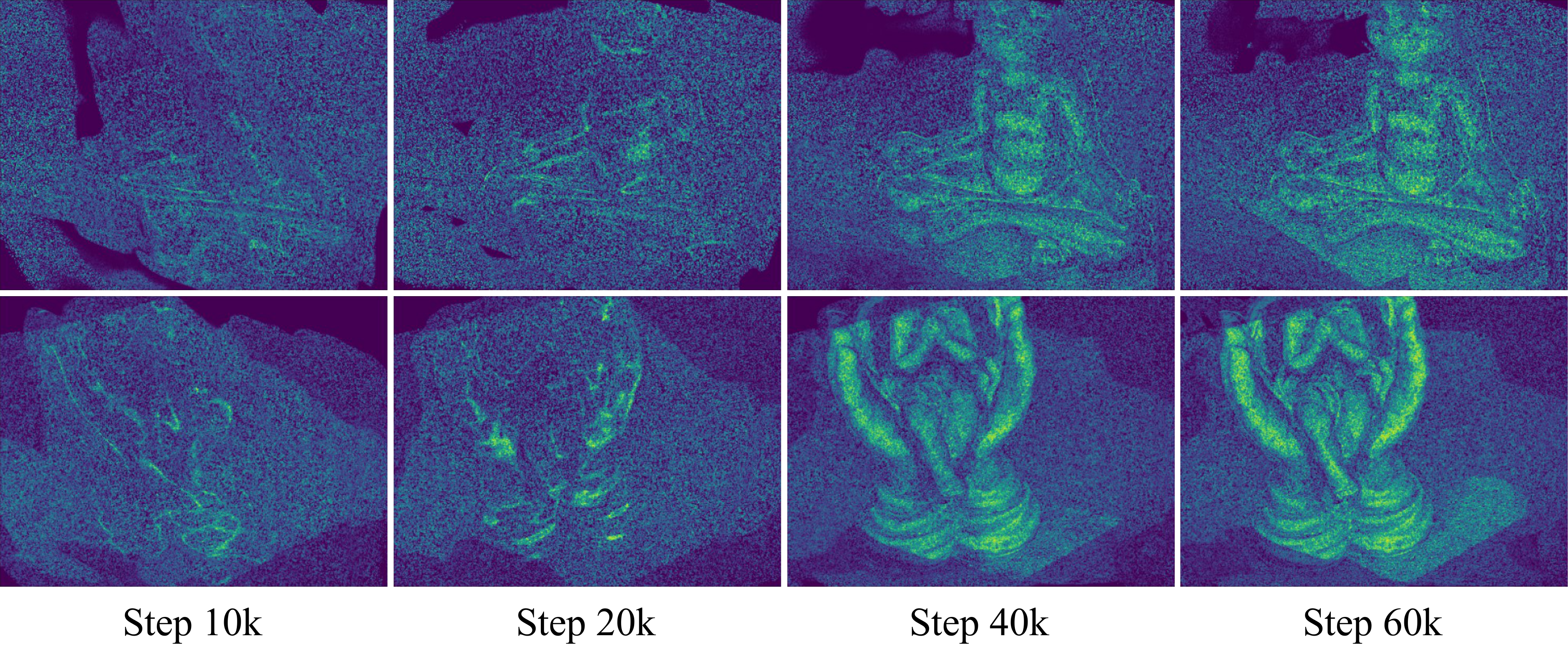}
    \caption{The variation of weighted feature similarity during training, brighter colors indicate higher feature similarity.}
    \label{fig:feat_sim_map}
\end{figure}

\begin{figure}[htb]
    \centering
    \includegraphics[width=1.0\linewidth]{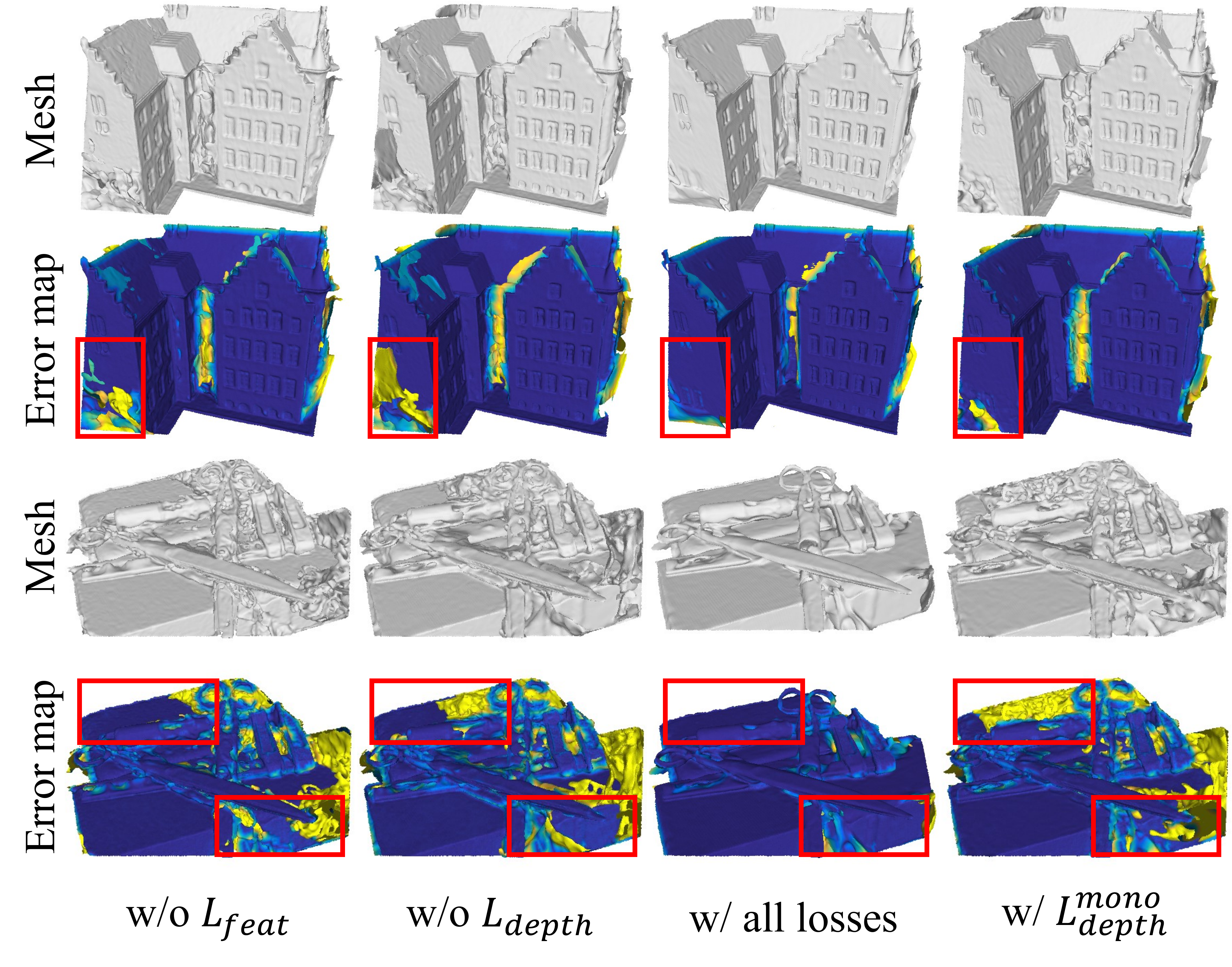}
    \caption{Visualization of reconstruction and error maps for scene scan24 and scan37 in DTU dataset with different losses. The differences of error maps are highlighted.}
    \label{fig:error_map}
\end{figure}

\subsection{Ablation Study}
We evaluate the components of our method with 3 small-overlapping views by an ablation study on the DTU \cite{jensen2014dtu} dataset. 
To compare the depth loss $L_{depth}^{mono}$ calibrated by rendered depth in MonoSDF \cite{yu2022monosdf}  with our depth loss $L_{depth}$, we replace $L_{depth}$ with $L_{depth}^{mono}$ to evaluate it in our method. 
We also compare the volume rendering-based feature consistency loss calculated using L1 distance (denoted as $L_{feat}^{L1}$) and L2 distance (denoted as $L_{feat}^{L2}$) with our method using feature similarity distance. We found that feature similarity distance is better than both L1 and L2 distance, as shown in Table \ref{tab:ablation}.

In addition, we replace our volume rendering-based feature consistency loss $L_{feat}$ with the on-surface feature consistency loss $L_{feat}^{surf}$ used in MVSDF \cite{zhang2021mvsdf} to compare the effectiveness of two different loss functions. Figure \ref{fig:surf_feat_and_ray_feat} illustrates the reconstruction results and error maps on the DTU dataset when using different feature consistency losses, under the on-surface feature consistency loss $L_{feat}^{surf}$, the meshes show large artifacts. 

Table~\ref{tab:ablation} shows the average Chamfer Distance over all 11 scenes on DTU dataset using different losses. The experimental results indicate that both feature consistency loss and uncertainty-guided depth constraint improve the surface reconstruction.

Figure~\ref{fig:feat_sim_map} illustrates the variation of the weighted feature similarity map during the training process.
Brighter colors indicate higher feature similarity, demonstrating that our volume rendering-based feature consistency loss can provide effective constraints.

Figure~\ref{fig:error_map} shows the reconstructed meshes and error maps for scene scan24 and scan37 on the DTU \cite{jensen2014dtu} dataset when using different losses. 
It can be observed that the mesh deteriorates without the volume rendering-based feature consistency loss or the uncertainty-guided depth constraint loss, and the reconstruction quality drops when using the depth loss $L_{depth}^{mono}$ in MonoSDF \cite{yu2022monosdf}.

\begin{figure}
    \centering
    \includegraphics[width=1.0\linewidth]{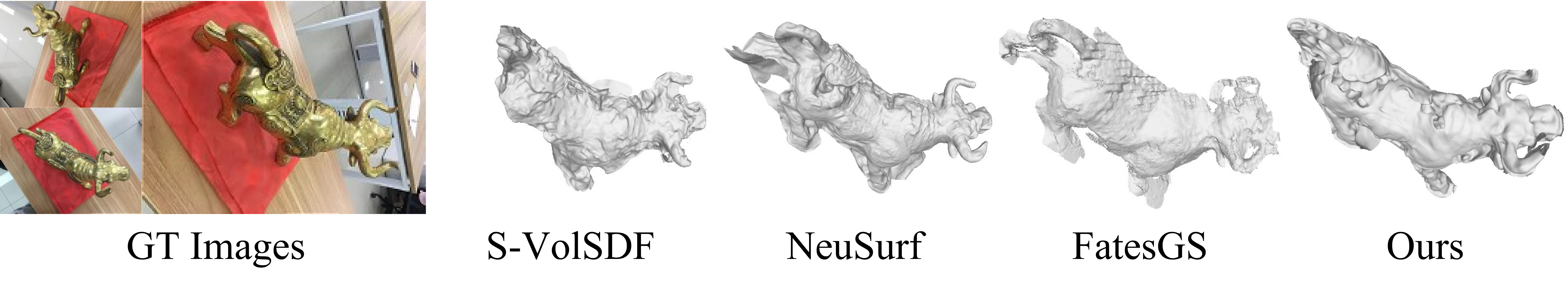}
    \caption{Failure case. For specular objects, the ambiguity in the color consistency constraint may lead to a rough surface.}
    \label{fig:failure_case}
\end{figure}

\section{Conclusions}
We propose a novel method for learning implicit representations from sparse views with small overlaps. Our novelty lies in a novel volume rendering-based feature consistency loss and an uncertainty-guided depth constraint.
Extensive experiments on the DTU \cite{jensen2014dtu} and BlendedMVS \cite{yao2020blendedmvs} datasets show that our method surpasses existing state-of-the-art sparse-view reconstruction methods in terms of reconstruction quality.

\textbf{Limitations.} Although our method shows significant improvement over other sparse view reconstruction methods, there are still some limitations. 
Firstly, for specular objects, the ambiguity in the color consistency constraint may lead to a rough surface, as shown in Figure \ref{fig:failure_case}. Secondly, Following previous studies \cite{wu2023svolsdf, huang2023neusurf, long2022sparseneus}, the camera poses of sparse views are obtained from the training dataset. However, in some cases, it may not be possible to obtain accurate camera poses using SfM methods like COLMAP \cite{schoenberger2016colmap} due to the lack of texture in the images or excessive viewing angles. Additionally, the feature consistency constraint method requires a pre-trained network to extract image features. The accuracy of the features determines the performance of the feature consistency constraint. 

\section{Acknowledgment}
The corresponding authors are Yu-Shen Liu and Haichuan Song. This work was supported by National Key R\&D Program of China (2022YFC3800600), and the National Natural Science Foundation of China (62272263), and in part by Tsinghua-Kuaishou Institute of Future Media Data.

{
    \small
    \bibliographystyle{ieeenat_fullname}
    \bibliography{main}
}


\end{document}